\documentclass[runningheads]{llncs}

\usepackage{eccv}

\usepackage{eccvabbrv}

\usepackage{graphicx}
\usepackage{booktabs}

\usepackage[accsupp]{axessibility}  

\usepackage[pagebackref,breaklinks,colorlinks,citecolor=eccvblue]{hyperref}

\usepackage{orcidlink}

\begin{document}

\title{DogWeave: High-Fidelity 3D Canine Reconstruction from a Single Image via Normal Fusion and Conditional Inpainting} 

\titlerunning{DogWeave}

\author{
Shufan Sun\textsuperscript{\textdagger *} \and
Chenchen Wang\textsuperscript{*} \and
Zongfu Yu
}

\authorrunning{S. Sun et al.}

\institute{
University of Wisconsin--Madison, Madison, WI 53706, USA\\
\email{ssun329@wisc.edu, cwang922@wisc.edu, zyu54@wisc.edu}\\
\textsuperscript{*} Equal contribution \quad
\textsuperscript{\textdagger} Corresponding author
}

\maketitle

\begin{center}
\includegraphics[width=1.0\textwidth]{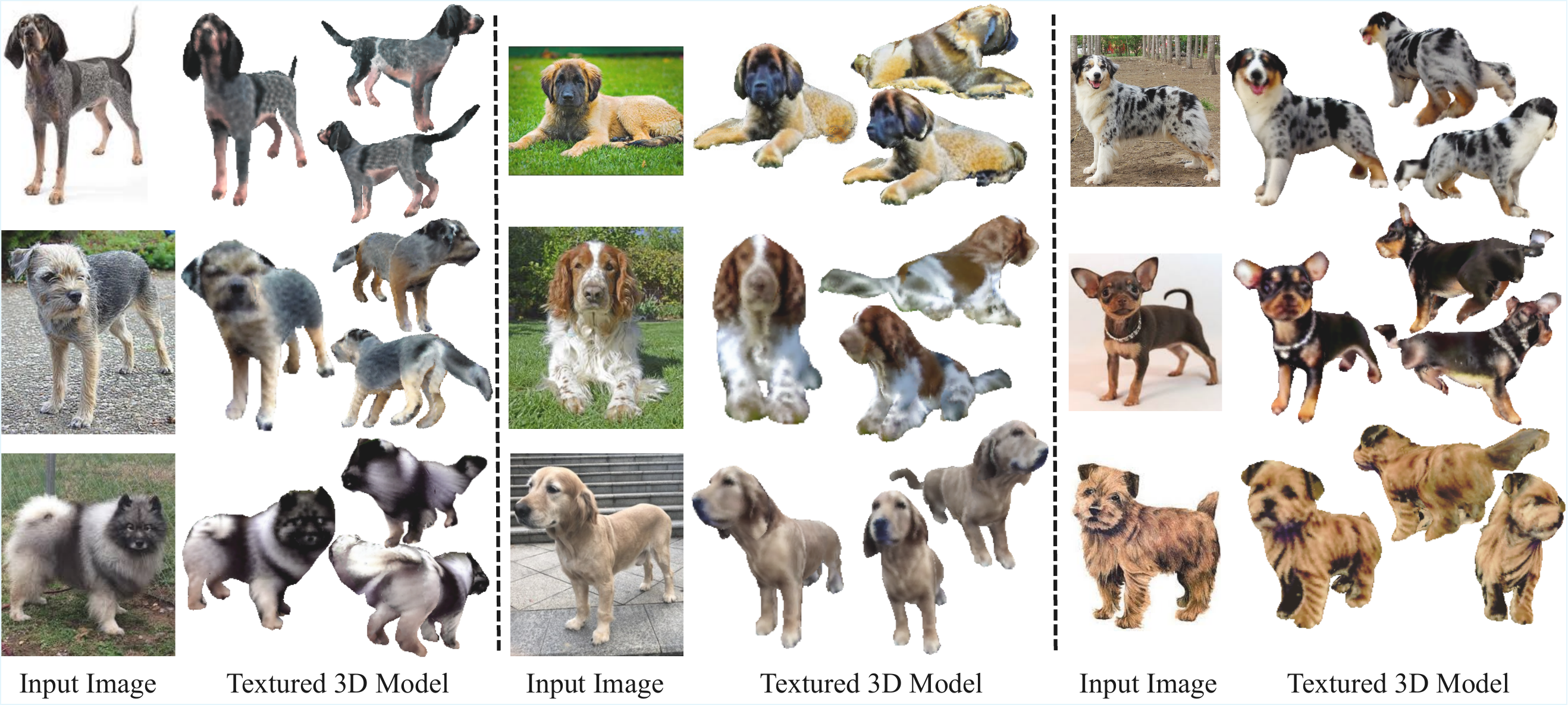}
\captionof{figure}{\textbf{DogWeave} reconstructs photorealistic 3D dog 
models from single images, achieving state-of-the-art texture fidelity 
and geometry-texture coherence.}
\label{fig:teaser}
\end{center}

\begin{abstract}

Monocular 3D animal reconstruction is challenging due to complex articulation, self-occlusion, and fine-scale details such as fur. Existing methods often produce distorted geometry and inconsistent textures due to the lack of articulated 3D supervision and limited availability of back-view images in 2D datasets, which makes reconstructing unobserved regions particularly difficult. To address these limitations, we propose \textbf{DogWeave}, a model-based framework for reconstructing high-fidelity 3D canine models from a single RGB image. DogWeave improves geometry by refining a coarsely-initiated parametric mesh into a detailed SDF representation through multi-view normal field optimization using diffusion-enhanced normals. It then generates view-consistent textures through conditional partial inpainting guided by structure and style cues, enabling realistic reconstruction of unobserved regions. Using only about 7,000 dog images processed via our 2D pipeline for training, DogWeave produces complete, realistic 3D models and outperforms state-of-the-art single image to 3d reconstruction methods in both shape accuracy and texture realism for canines.
\end{abstract}

\section{Introduction}
\label{sec:intro}

High-quality 3D animal models have broad applications in animation and game production. Realism requires not only accurate geometry but also vivid textures that capture fine-grained details such as fur patterns, wrinkles, and surface variations. Even small differences in fur density, color distribution, or shading can drastically affect perceived identity. Traditional pipelines rely on UV painting\cite{miller2022color,uvpainting,algasov2025unsupervised}, physically based fur simulations\cite{kajiya1989rendering, sklyarova2026neuralfur,yan2015physically, xu2025enhanced,sklyarova_kabadayi_2025neuralfur}, or large-scale 3D scanning systems\cite{medina2024picocam,yang2025synchronizing,omotara2024high,Irschick2022}, requiring specialized hardware and multi-view captures and making high-fidelity reconstruction costly and difficult to scale.

Automating 3D quadruped reconstruction from a single image aims to reduce the need for complex data-capture setups. Existing methods face a fundamental data–geometry–texture trilemma. \textbf{Few-Shot Fast Inference} methods \cite{wang2024crm,trellis,sam3dteam2025sam3d3dfyimages,long2023wonder3d,TripoSR2024,li2024craftsman,yang2024hunyuan3d,hunyuan3d22025tencent,lai2025hunyuan3d25highfidelity3d}, which rely on large-scale 3D datasets or controlled multi-view captures \cite{objaverse,objaverseXL,khanna2023hssd,stojanov2021shape,fu20213dfuture} to directly infer 3D representations, offer efficiency and flexibility. However, they lack strong anatomical priors and must predict textures for unobserved regions based solely on training statistics. Due to limited dataset specificity and domain coverage, these methods often produce over-smoothed textures, drifted color distributions, and incorrect topology. Given that most publicly available animal datasets \cite{KhoslaJayadevaprakashYaoFeiFei_FGVC2011,Zou2020ThuDogs,Cermak_2024_WACV,deng2009imagenet,Lei_2024_CVPR} contain only single-view images—typically frontal or lateral and organized by breed—\textbf{Model-Based} approaches \cite{dogrecon2024,Sabathier2024AnimalAvatars,rueegg2023bite,li2024fauna,zuffi2017smal} were introduced to leverage abundant 2D supervision. While this reduces dependence on 3D or multi-view data, single-view supervision introduces inherent shape ambiguities, often resulting in low-fidelity geometry and misaligned textures. Consequently, current reconstruction methods struggle with two critical limitations: \textit{geometric fidelity} and \textit{texture similarity}. Reconstructed models frequently fail to maintain photorealistic and consistent identity across novel viewpoints, particularly in unobserved regions where both geometric and appearance ambiguities are the highest.

We present \textbf{DogWeave}, a \textit{model-based} optimization framework for photorealistic, identity-consistent 3D dog reconstruction from a single image. Our key insight is that canine appearances cluster by breed and fur patterns, making style-controlled texturing a natural source of identity guidance. DogWeave first refines a parametric mesh using perceptual losses, converts it to a signed distance field (SDF) for stable geometric manipulation, and then applies diffusion-enhanced multi-view normal fusion to recover fine-scale surface details. A sequential geometry- and style-conditioned inpainting scheme progressively textures each viewpoint, enforcing view-consistent, high-frequency textures through volumetric geometry–texture coupling. Trained using only 2D images without ground-truth 3D data, DogWeave outperforms all baselines, achieving superior photorealism and identity consistency in novel-view synthesis for canines.

\paragraph{\textbf{Contributions}.}
\begin{itemize}
\item We introduce a \textit{model-based single-image-to-3D reconstruction framework} for high-fidelity textured canine models, trained solely with single-view images and without ground-truth 3D supervision.
\item We propose a \textit{sequential geometry- and style-conditioned inpainting} scheme that textures each view while preserving \textit{identity-specific} appearance.
\item We perform a \textit{comprehensive evaluation} showing superior geometry–texture coherence and photorealism over existing methods.
\end{itemize}

\section{Related Works}
\subsection{Single-Image to 3D Reconstruction for Animals}

\textbf{Parametric models} \cite{zuffi2017smal,rueegg2023bite} provide statistical priors for animal shape and articulation. \textit{SMAL}~\cite{zuffi2017smal} models shape variability via learned components and skeletal rigging, while \textit{BITE}~\cite{rueegg2023bite} introduces D-SMAL, a dog-specific variant trained on 3D scans of multiple breeds. These models, however, produce low-fidelity geometry with limited fine-scale details such as imprecise facial boundaries, causing geometry–texture misalignment in downsteam reconstruction tasks.

\textbf{Model-based}\cite{Sabathier2024AnimalAvatars,dogrecon2024,li2024fauna} methods refine parametric models through optimization or learned deformation. \textit{Animal Avatars}~\cite{Sabathier2024AnimalAvatars} fits an enhanced SMAL model with dense surface embeddings and implicit texture to monocular videos. Despite improved alignment, per-frame fitting can produce misaligned features and color discontinuities. \textit{DogRecon}~\cite{dogrecon2024} reconstructs 3D animatable dogs from a single image by combining D‐SMAL-derived canine priors with canine‑centric novel view synthesis and Gaussian splatting, but independent view synthesis and implicit optimization can introduce inconsistent texture and residual geometric noise across views. \textit{3D-Fauna}~\cite{li2024fauna} learns a pan‑category deformable 3D animal model for over 100 quadruped species using 2D internet images, achieving broad generalization but emphasizes coarse articulated shape and category‑level modeling rather than object‑specific fine geometry or detailed texturing.

\textbf{Few-Shot Fast Inference}~\cite{TripoSR2024,wang2024crm,trellis,sam3dteam2025sam3d3dfyimages,long2023wonder3d,li2024craftsman,hunyuan3d22025tencent,lai2025hunyuan3d25highfidelity3d,yang2024hunyuan3d} methods regress 3D representations from a single image using feed-forward networks trained on large synthetic corpora~\cite{objaverse,objaverseXL,khanna2023hssd,stojanov2021shape,fu20213dfuture}. 
\textit{TripoSR}~\cite{TripoSR2024} and \textit{CRM}~\cite{wang2024crm} achieve real-time reconstruction through triplane regression, but lack anatomical priors and tend to produce over-smoothed geometry with washed-out textures for quadrupeds. 
\textit{TRELLIS}~\cite{trellis} encodes geometry and appearance through sparse structured 3D latents trained on large-scale 3D datasets, yet without explicit identity grounding it struggles to recover distinctive fur patterns and markings.
Similarly, \textit{SAM3D}~\cite{sam3dteam2025sam3d3dfyimages} leverages large-scale segmentation priors for object-centric reconstruction, but its geometry and texture predictions remain conditioned on generic visual features and often miss fine-scale appearance details.
\textit{Hunyuan3D}~\cite{yang2024hunyuan3d,hunyuan3d22025tencent,lai2025hunyuan3d25highfidelity3d} scales diffusion-based multi-view generation for 3D asset synthesis, but like other category-agnostic models, it lacks strong identity grounding and struggles to preserve consistent fine-scale details from a single input image.

Moving beyond direct regression, \textit{Wonder3D}~\cite{long2023wonder3d} improves surface detail via cross-domain diffusion over multi-view normals and color; however, it relies on diffusion-based multi-view synthesis, which may introduce inconsistent geometry or textures in unobserved regions, resulting in identity drift for articulated quadrupeds.
\textit{CraftsMan3D}~\cite{li2024craftsman} prioritizes geometric fidelity through cross-view attention, yet limited canine-specific supervision and quadruped depth ambiguities often produces shapes influenced by dataset priors rather than preserving instance-specific identity.

\subsection{Conditioning Diffusion Models for Structural Control}

Latent diffusion models~\cite{rombach2022stablediffusion} achieve high-resolution image synthesis through iterative denoising in compressed latent spaces and have been extended beyond generation. \textit{ControlNet}~\cite{zhang2023controlnet} enables structural conditioning with trainable encoders for inputs like depth or normals, while \textit{IP-Adapter}~\cite{ye2023ipadapter} injects appearance references via decoupled cross-attention for identity-preserving style transfer. Diffusion priors have also been adapted for geometry: \textit{Marigold}~\cite{ke2023repurposing,ke2025marigold} fine-tunes a generative diffusion model to predict monocular depth and normals, recovering fine-scale details, though it can underestimate depth in occluded quadruped regions. \textit{HumanNorm}~\cite{huang2024humannorm} further leverages predicted normal and depth maps as structural guidance for text-to-3D human generation.

\subsection{Volumetric Representations and Texturing for 3D Avatars}

\textit{Signed distance fields} (SDFs)~\cite{Wang-Sig2022,Park_2019_CVPR,yariv2021volume,lorensen1987marching} represent surfaces as zero level-sets, enabling exact normal computation and watertight geometry. \textit{HumanRef}~\cite{zhang2024humanref} demonstrates that SDF grids can be refined by rendering multi-view projections, enhancing them via conditional diffusion, and backpropagating corrections onto the grid. However, the scarcity of multi-view 3D canine scans precludes reliable normal estimation, undermining the geometric constraints that stabilize SDS-guided texture synthesis~\cite{huang2024humannorm,zhang2024humanref} in unobserved regions.

To address texturing in occluded regions, \textit{progressive inpainting} approaches~\cite{albahar2023sgd,perla2024easitex} sequentially backproject 2D diffusion-generated textures onto meshes view-by-view, preserving geometry–texture coherence through iterative refinement. \textit{Human SGD}~\cite{albahar2023sgd} relies on strong shape priors and large-scale multi-view supervision unavailable for canines, limiting identity consistency in unobserved regions. \textit{Easitex}~\cite{perla2024easitex} employs edge-aware partial inpainting to improve boundary sharpness, but its dependence on well-defined mesh edges makes it ill-suited for dogs, whose smooth fur surfaces provide no reliable structural boundaries. \textit{DogWeave} addresses these limitations by coupling SDF-based geometry with sequential identity-conditioned inpainting, enforcing appearance consistency in unobserved regions without requiring 3D supervision or sharp geometric boundaries.

\section{Methods}

\begin{figure*}[t!]
    \centering
    \includegraphics[width=\textwidth]{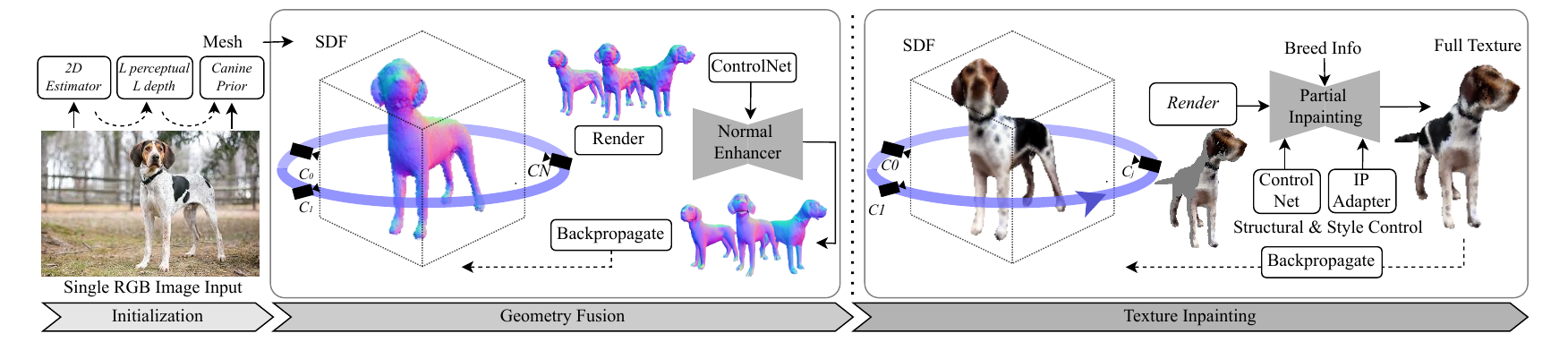}
    \caption{
\textbf{DogWeave} reconstructs high-fidelity 3D dog models from a single RGB image using a three-stage pipeline. \textit{Stage 1: Coarse Shape Initialization} generates a base mesh with BITE~\cite{rueegg2023bite} and refines overall proportions. \textit{Stage 2: Surface Detail Enhancement} converts the mesh to a volumetric SDF and incorporates diffusion-enhanced multi-view normals for fine geometric details. \textit{Stage 3: Sequential Texturing} produces photorealistic, identity-consistent appearance through style- and breed-conditioned inpainting.
}
    \label{fig:pipeline}
\end{figure*}

Figure~\ref{fig:pipeline} shows our three-stage pipeline for reconstructing high-fidelity 3D dog models from single images, with details described below. 

\subsection{Coarse Geometry Initialization}
\label{sec:coarse_init}

\begin{figure}[h!]  
    \centering
    \includegraphics[width=\linewidth]{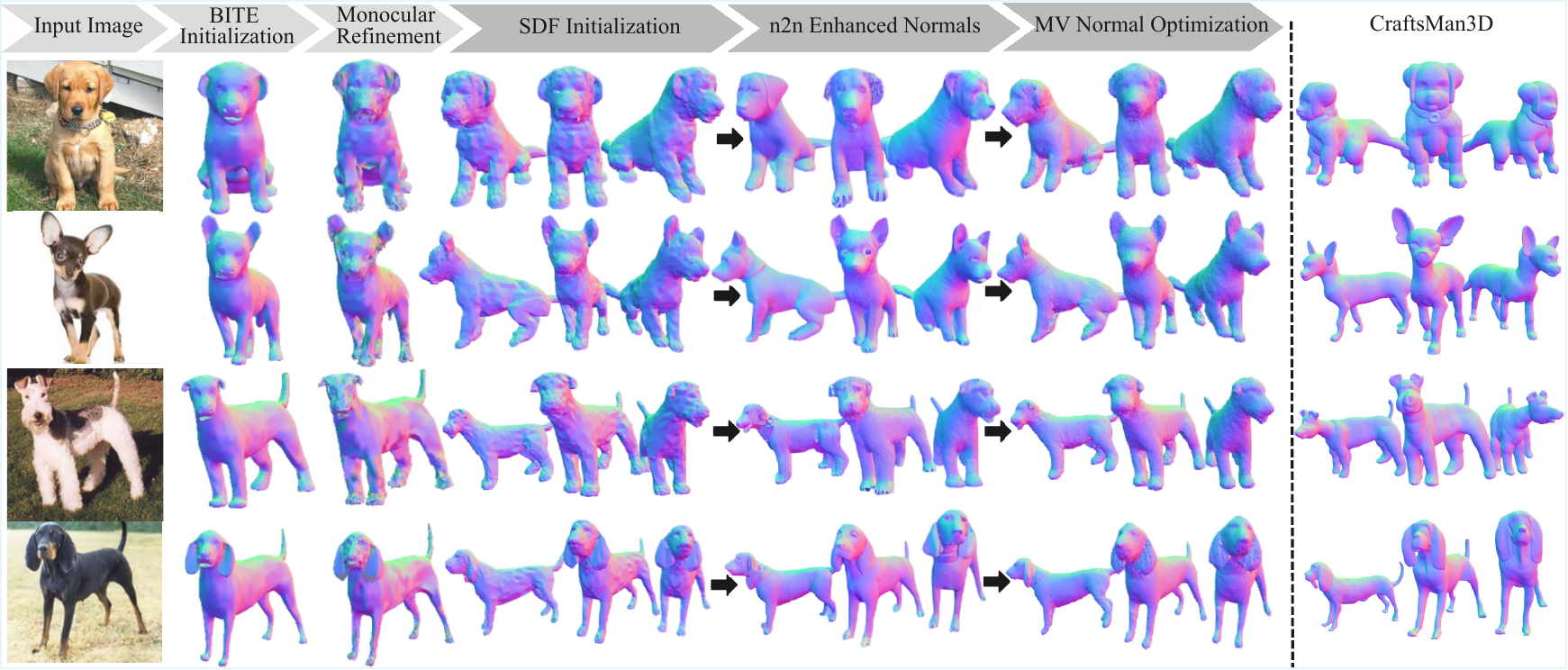}
    \caption{\textbf{Normal optimization.} We progressively refine geometry from the BITE base mesh (light arrows) and through SDF optimization with multi-view normal fusion (dark arrows) to recover fine-scale surface details. The right column compares to CraftsMan3D\cite{li2024craftsman}, which exhibits domain drift when species constraints are not applied.}
    \label{fig:geometry_demo}
\end{figure}

After initializing the base geometry using BITE~\cite{rueegg2023bite}, we extend its optimization with monocular depth and normal supervision from Marigold~\cite{ke2023repurposing,ke2025marigold} (Figure~\ref{fig:geometry_demo}, lighter arrows), improving shape fitting beyond silhouette alignment. Nevertheless, BITE’s fixed topology limits its ability to recover fine-scale surface details such as wrinkles that remain consistent under varying illumination.

\textbf{Progressive refinement.}
We refine the mesh in two stages by minimizing:
\begin{equation}
\mathcal{L}_{\text{total}} = \lambda_d \mathcal{L}_{\text{depth}} + \lambda_n \mathcal{L}_{\text{normal}} + \lambda_s \mathcal{L}_{\text{silh}} + \lambda_r \mathcal{L}_{\text{smooth}}
\end{equation}

\textbf{Stage 1} emphasizes depth alignment to establish aligned body proportions. \textbf{Stage 2} shifts to perceptual normal matching to recover surface features.

Our \textbf{perceptual normal loss}\cite{otto2023perceptual} uses VGG-16 features\cite{johnson2016perceptual} at three layers:
\begin{equation}
\mathcal{L}_{\text{normal}} = \sum_{\ell}\|f_\ell(\hat{N}) - f_\ell(N)\|_2^2 + 2\sum_{\ell}\|G_\ell(\hat{N}) - G_\ell(N)\|_F^2
\end{equation}
where $\hat{N}/N$ are rendered/target normals; $f_\ell(\cdot)$ extracts features from the $\ell$-th VGG-16 layer, and $G_\ell(\cdot) = \frac{1}{chw}FF^T$ is the normalized Gram matrix\cite{gatys2016image}, which captures cross-channel feature 
correlations to enforce consistent local surface \textit{style} independent 
of spatial pixel alignment. The content term enforces geometric fidelity 
while the style term regularizes local surface patterns, 
preserving perceptually meaningful curvature without requiring 
per-pixel correspondence.

Our \textbf{scale-invariant depth loss}\cite{eigen2014depth} handles monocular ambiguity:
\begin{equation}
\mathcal{L}_{\text{depth}} = \tfrac{1}{n}\sum_{i=1}^{n} d_i^2 - 0.5\left(\tfrac{1}{n}\sum_{i=1}^{n} d_i\right)^2, \quad d_i = \log \hat{D}_i - \log D_i
\end{equation}
where $\hat{D}_i/D_i$ are rendered/target depths. The variance-based formulation optimizes relative depth structure while accommodating scale shifts and biases in monocular predictions. Standard regularization ($\mathcal{L}_{\text{silh}}$, $\mathcal{L}_{\text{smooth}}$) follows BITE~\cite{rueegg2023bite}.

\subsection{SDF-Based Geometric Refinement}
\label{sec:sdf_refine}

To recover fine surface details while preserving domain-specific cues (e.g., facial features) and mitigating depth ambiguities from Marigold’s monocular normals~\cite{ke2025marigold}, we fine-tune a ControlNet~\cite{zhang2023controlnet}-based diffusion model on Stable Diffusion 1.5~\cite{rombach2022stablediffusion} using our paired dataset (Section~\ref{sec:data_prep}), rather than using the training-free generic enhancer in CraftsMan3D\cite{li2024craftsman} (Figure~\ref{fig:geometry_demo}). Since Marigold’s estimates are unaligned in 3D, they are unreliable for multiview cross-attention training. We therefore adopt a view-weighted fusion strategy that prioritizes field-specific fidelity over strict multiview consistency.

We convert the coarse mesh into a $256^3$ volumetric SDF~\cite{wang2022dualocnn,lorensen1987marching} and initialize a $256^3$ normal field from uniformly sampled azimuth views. For each view $i$, we (1) render camera-space normals $\mathbf{n}_i^c$ via sphere tracing, (2) enhance them with our diffusion model, (3) transform to world space $\mathbf{n}_i^w = \mathbf{R}_i^T \mathbf{n}_i^c$, and (4) accumulate into the normal field, resolving overlaps via weighted averaging:

\begin{equation}
\mathbf{N}_{\text{target}}(\mathbf{x}) = \frac{\sum_{i \in V(\mathbf{x})} w_i(\mathbf{x}) \mathbf{n}_i^w(\mathbf{x})}{\left\|\sum_{i \in V(\mathbf{x})} w_i(\mathbf{x}) \mathbf{n}_i^w(\mathbf{x})\right\|}
\end{equation}

Here, $V(\mathbf{x})$ is the set of views whose rays intersect voxel $\mathbf{x}$, and $w_i(\mathbf{x})$ counts the rays hitting that voxel as a confidence weight. L2-normalization ensures the fused normal is a unit vector. This weighted fusion integrates multi-view information while reducing noise and preserving fine-scale details.

\textbf{SDF Optimization.} We optimize SDF values to match the fused target normals via a volumetric consistency loss. For each voxel $\mathbf{x} \in \mathcal{S}$, where $\mathcal{S} = \{\mathbf{x} \mid \sum_i w_i(\mathbf{x}) > 0\}$ denotes voxels with at least one valid projection, we minimize the cosine distance between the SDF gradient and the fused target:

\begin{equation}
    \mathcal{L}_{\text{normal}} = \mathbb{E}_{\mathbf{x} \in \mathcal{S}}\left[1 - \mathbf{n}(\mathbf{x}) \cdot \mathbf{N}_{\text{target}}(\mathbf{x})\right]
\end{equation}

Here, $\mathbf{n}(\mathbf{x}) = \nabla \phi(\mathbf{x}) / \|\nabla \phi(\mathbf{x})\|$ is the SDF gradient normal at voxel $\mathbf{x}$. To resolve depth ambiguities from 2D normal maps and preserve global shape, we add a per-view mask loss using binary cross-entropy between sphere-traced occupancy masks $\hat{M}_i$ and target masks $M_i$, along with an eikonal penalty
\begin{equation}
    \mathcal{L}_{\text{eikonal}} = \mathbb{E}_{\mathbf{x}}\!\left[\left(\|\nabla \phi(\mathbf{x})\| - 1\right)^2\right],
\end{equation}
which enforces the defining property of a signed distance field and prevents gradient drift in voxels outside $\mathcal{S}$. The full objective is:
\begin{equation}
    \mathcal{L} = \lambda_n \mathcal{L}_{\text{normal}} + \lambda_m \mathcal{L}_{\text{mask}} + \lambda_e \mathcal{L}_{\text{eikonal}}.
\end{equation}

\subsection{Geometry-Aware Texture Synthesis} 
\label{sec:texture}

\begin{figure}[h!]  
    \centering
    \includegraphics[width=\linewidth]{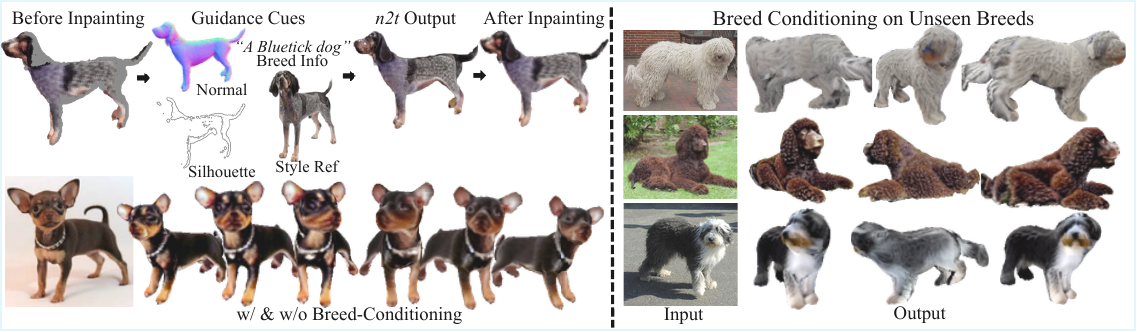}
    \caption{\textbf{Breed Information Visualizations.} Upper left: breed information guides the texturing pipeline. Lower left: improved facial feature precision when breed info for this Toy Terrier is included. Right: robust generalization to unseen breeds.}
    \label{fig:breed_info}
\end{figure}

We synthesize photorealistic, multi-view consistent textures by progressively completing a volumetric color field through geometry-conditioned inpainting (Figure~\ref{fig:pipeline}, right; Figure~\ref{fig:breed_info}, upper left).

\textbf{Color field initialization.} We initialize a $256^3$ color grid on the SDF, where each voxel stores RGB color $\mathbf{c} \in [0,1]^3$ and a confidence weight $w \in \mathbb{R}_+$. Surface normals rendered from the optimized SDF provide geometric conditioning.

\textbf{Normal-conditioned inpainting.} We train a conditional inpainting model on Stable Diffusion 1.5~\cite{rombach2022stablediffusion} with IP-Adapter~\cite{ye2023ipadapter}. The model takes five inputs: (1) rendered normals, (2) current partial texture, (3) mask of incomplete regions, (4) input image as style reference, and (5) breed information. Training data is described in Section~\ref{sec:data_prep}.

\textbf{Sequential projection.} Views from a circular camera array are processed in spiral order to maximize angular coverage. For each view, we (1) render the current texture to identify incomplete regions, (2) generate completed textures via diffusion inpainting, and (3) project colors onto the volumetric grid.

\textbf{Volumetric projection with confidence weighting.} Instead of projecting isolated points, each ray-surface intersection colors a small cylindrical neighborhood of radius $r=1$ voxel. For an intersection at point $\mathbf{p}$, we sample neighboring voxel positions $\mathbf{x}$ with depth offset $\delta_d \in [-r, r]$ along the ray and radial distance $\delta_r \in [0, r]$ perpendicular to it. Each voxel receives a spatial blending weight:

\begin{equation}
w_{\text{blend}}(\delta_d, \delta_r) = \left(1 - \frac{|\delta_d|}{r+1}\right)\left(1 - \frac{\delta_r}{r+1}\right),
\end{equation}

which decays linearly from 1 at the intersection to 0 at radius $r$. A view-dependent confidence weight for view $i$ is:

\begin{equation}
w_{\text{conf}}(\mathbf{x}, i) = |\mathbf{n}(\mathbf{x}) \cdot \mathbf{v}_i| \cdot \frac{\beta_i}{1 + \|\mathbf{x} - \mathbf{c}_i\|},
\end{equation}

where $\mathbf{n}(\mathbf{x})$ is the surface normal, $\mathbf{v}_i$ and $\mathbf{c}_i$ are the view direction and camera position, and $\beta_i$ prioritizes diffusion-enhanced views. The final voxel weight is $w(\mathbf{x}) = w_{\text{blend}} \cdot w_{\text{conf}}$. We project only to uncolored voxels with low confidence to preserve texture continuity, and remaining gaps are filled via iterative neighborhood averaging.

\section{Experiments}
\label{sec:experiments}

We evaluate DogWeave on single-image 3D dog reconstruction, comparing against state-of-the-art methods in geometric accuracy, texture quality, and multi-view consistency. Results show superior recovery of fine surface details while maintaining photorealistic appearance across viewpoints.

\subsection{Self-Supervised Training Data Construction}
\label{sec:data_prep}

Without 3D scans or multi-view data, we construct paired training data self-supervised from single-view images.

\textbf{Dataset.} We use $\sim$7K images from the Tsinghua Dog Dataset~\cite{Zou2020ThuDogs}, covering diverse breeds, poses, and lighting.

\textbf{Normal pairs.} For each image, we optimize a BITE mesh with Marigold depth and normal supervision (Section~\ref{sec:coarse_init}) and render normals from the optimized mesh. Paired with Marigold-predicted normals~\cite{ke2025marigold} aligned via silhouette correspondence, this yields $(\mathbf{N}_{\text{coarse}}, \mathbf{N}_{\text{fine}})$ pairs.

\textbf{Normal-texture pairs.} Marigold-predicted normals are paired with object RGB images segmented via silhouettes to form $(\mathbf{N}_{\text{fine}}, \mathbf{T})$ training pairs.

\subsection{Implementation Details}
\textbf{Diffusion model training.} We use the Stable Diffusion 1.5 ~\cite{rombach2022stablediffusion} with ControlNet~\cite{zhang2023controlnet} and train only the ControlNet branches while keeping the Stable Diffusion backbone frozen for both tasks. 
For normal enhancement, we train the model for 50 epochs using batch size~8, learning rate $10^{-4}$, AdamW\cite{loshchilov2017decoupled} optimization, and cosine annealing. The training data consist of the normal pairs $(\mathbf{N}_{\text{coarse}}, \mathbf{N}_{\text{fine}})$ described in Section~\ref{sec:data_prep}. For partial texture inpainting, we adopt the inpainting version of Stable Diffusion 1.5, conditioned on geometry via ControNet and on appearance via IP-Adapter (16 tokens, scale~1.0). This model is trained for 80 epochs using the normal–texture pairs $(\mathbf{N}_{\text{fine}}, \mathbf{T})$ described in Section~\ref{sec:data_prep}. During training, we additionally apply erosion-based random masks to the textured RGB patches to simulate view-dependent missing regions. All other hyperparameters follow the normal-enhancement setup. Training all models requires approximately 12 hours on a single NVIDIA~A100 GPU.

\textbf{Coarse mesh optimization.} We refine BITE geometry in two 100-iteration stages using Adam~\cite{Kingma2014AdamAM}. Stage 1 emphasizes depth ($\lambda_d=1.0$, $\lambda_n=0.5$, LR $=10^{-3}$), while Stage 2 emphasizes normals ($\lambda_d=0.0$, $\lambda_n=1.0$, LR $=10^{-4}$).

\textbf{SDF refinement.} We optimize a $256^3$ SDF grid for 50 iterations using Adam (LR $=2\!\times\!10^{-4}$). Multi-view normals are accumulated from 8 evenly spaced views at 45° intervals and projected into a $256^3$ normal field via ray marching (256 samples per ray). Unfilled surface voxels are filled via 6-connected neighborhood averaging for 5 iterations. The normal consistency loss weight is 0.3.

\textbf{Texture synthesis.} For $K=8$ spiral views, we (1) render partial textures (512 ray samples), (2) inpaint incomplete regions (50 DDIM steps, guidance 7.5), and (3) project via volumetric tubes ($r=1$ voxel, confidence threshold 0.03). 

\textbf{Inference.} Given a single RGB image, coarse mesh refinement takes $\sim$1 min, SDF generation and optimization $\sim$4 min (50 iterations), and texture completion $\sim$3 min (8 views), totaling $\sim$8 min on a single A100 GPU.

\textbf{Baselines.} We compare against Wonder3D~\cite{long2023wonder3d}, TRELLIS~\cite{trellis}, CRM~\cite{wang2024crm}, TripoSR~\cite{TripoSR2024}, Fauna~\cite{li2024fauna}, SAM3D~\cite{sam3dteam2025sam3d3dfyimages}, and Hunyuan3D\cite{hunyuan3d22025tencent,yang2024hunyuan3d,lai2025hunyuan3d25highfidelity3d} using official implementations with recommended settings.

\subsection{Evaluation Metrics}

\begin{figure*}[t!]
    \centering
    \includegraphics[width=\textwidth]{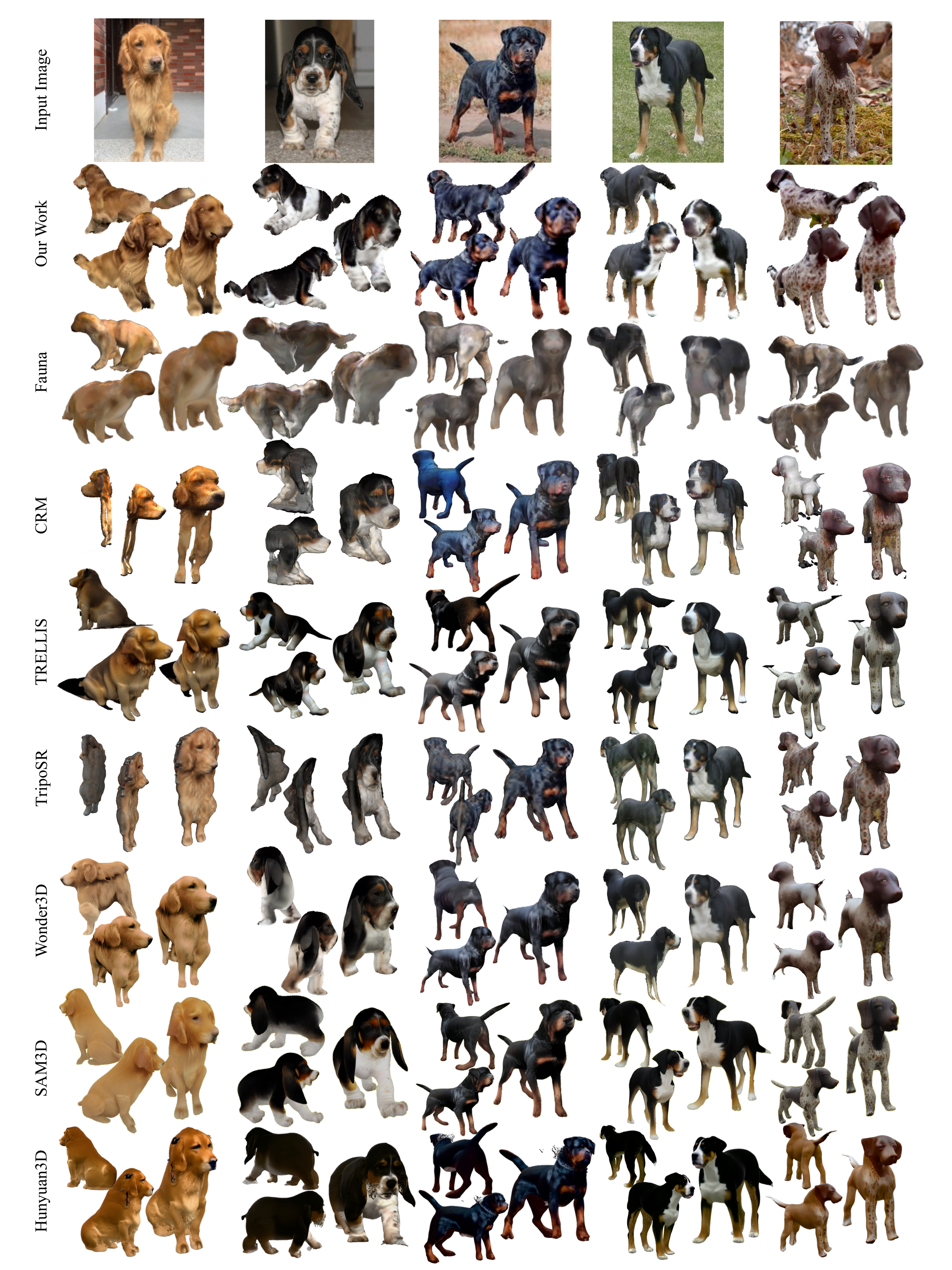}
\caption{
\textbf{Qualitative comparison across breeds and poses.}
\textit{Fauna}~\cite{li2024fauna} produces coarse textures.
\textit{CRM}~\cite{wang2024crm} and \textit{TripoSR}~\cite{TripoSR2024} show color drift and topology errors.
\textit{Wonder3D}~\cite{long2023wonder3d} exhibits geometry bias.
\textit{Trellis}~\cite{trellis} and \textit{SAM3D}~\cite{sam3dteam2025sam3d3dfyimages} show similar artifacts.
\textit{Hunyuan3D}~\cite{hunyuan3d22025tencent,yang2024hunyuan3d,lai2025hunyuan3d25highfidelity3d} generates good geometry but blurred textures in occluded regions.
}
    \label{fig:comparison}
\end{figure*}

\begin{table*}[ht]
\centering
\small
\caption{Evaluation of 3D Dog Reconstruction: Perceptual Similarity and Realism Metrics}
\label{tab:results}
\begin{tabular}{lcccc}
\hline
\textbf{Method / Metric} & \textbf{FID$ \downarrow$\cite{Seitzer2020FID}} & \textbf{CLIP $\uparrow$\cite{ilharco_gabriel_2021_5143773}} & \textbf{LPIPS $\downarrow$\cite{zhang2018perceptual}} & \textbf{DreamSim $\downarrow$\cite{fu2023dreamsim}} \\
\hline
Our Work & 176.4 & 0.9081 & 0.2495 & 0.1751 \\
Hunyuan3D & 194.3 & 0.8874 & 0.2813 & 0.1843 \\
SAM3D & 219.3 & 0.8491 & 0.3017 & 0.2394 \\
trellis & 235.7 & 0.8628 & 0.3223 & 0.2788 \\
Wonder3D  & 297.5 & 0.8199 & 0.3180 & 0.2620 \\
CRM  & 322.5 & 0.7270 & 0.2985 & 0.3811 \\
TripoSR & 384.6 & 0.6835 & 0.3202 & 0.4242 \\
Fauna & 393.5 & 0.6052 & 0.3963 & 0.4656 \\

\hline
\end{tabular}
\end{table*}

We evaluate on 50 randomly selected images from the Stanford Dog Dataset~\cite{KhoslaYaoJayadevaprakashFeiFei_FGVC2011} (cross-dataset from ThuDogs~\cite{Zou2020ThuDogs}), covering 45 breeds with diverse sizes, fur types, poses (42\% standing, 28\% sitting, 18\% lying, 12\% action), and viewpoints (48\% frontal, 32\% side, 20\% 3/4), excluding multi-object or heavily occluded cases. The full evaluation set will be publicly released on GitHub.

High-quality 3D dog scans are scarce, making geometry-level evaluation impractical at scale. We instead adopt perceptually aligned 2D metrics on multi-view renderings against the input RGB: \textbf{FID}~\cite{Seitzer2020FID} measures distributional realism; \textbf{CLIP Score}~\cite{ilharco_gabriel_2021_5143773} evaluates semantic breed identity; \textbf{LPIPS}~\cite{zhang2018perceptual} captures texture fidelity and fine details; and \textbf{DreamSim}~\cite{fu2023dreamsim} measures mid-level perceptual similarity aligned with human judgment. Lower FID, LPIPS, and DreamSim indicate better quality, while higher CLIP Score indicates stronger identity preservation. Values are averaged across views and subjects.

\textbf{Multi-view consistency.} We additionally assess texture seamlessness and geometric plausibility through visual inspection of rendered sequences (Figure~\ref{fig:comparison}).

\subsection{Quantitative Results}

Table~\ref{tab:results} compares methods on perceptual quality metrics. 

DogWeave outperforms all baselines across all perceptual metrics. 
It achieves the lowest FID (176.4), improving over the next-best method 
(Hunyuan3D, 194.3) by approximately 9\%, indicating stronger visual realism. 
Our method also attains the highest CLIP score (0.9081), suggesting better 
preservation of breed-specific semantic features. 

In addition, DogWeave yields the lowest LPIPS (0.2495) and DreamSim (0.1751), 
indicating improved perceptual similarity to the input images while 
maintaining consistent appearance across novel viewpoints.

\subsection{Qualitative Results}

Figure~\ref{fig:comparison} presents comparisons across breeds and poses. DogWeave consistently preserves coloring, fine details, and anatomical correctness across viewpoints. Compared to baselines, it maintains identity-specific fur patterns and handles challenging articulated poses, while other methods exhibit coarse textures, color drift, topology errors, or geometry bias.



\subsection{Ablation Studies}

\begin{table}[ht]
\centering
\small
\caption{Quantitative ablation of 3D Dog Reconstruction}
\label{tab:results2}
\begin{tabular}{lcccc}
\hline
\textbf{Method / Metric} & \textbf{LPIPS $\downarrow$} & \textbf{DreamSim $\downarrow$} \\
\hline
Our Work  & 0.2495 & 0.1751\\
w/o perceptual losses  & 0.2641 & 0.1821 \\
w/o normal enhancer   & 0.2661 & 0.1955 \\
w/o IP-Adapter  & 0.2713 & 0.2546 \\
w/o Breed Info & 0.2570 & 0.1949 \\
\hline
\end{tabular}
\end{table}
\begin{figure}[h!]  
    \centering
    \includegraphics[width=\linewidth]{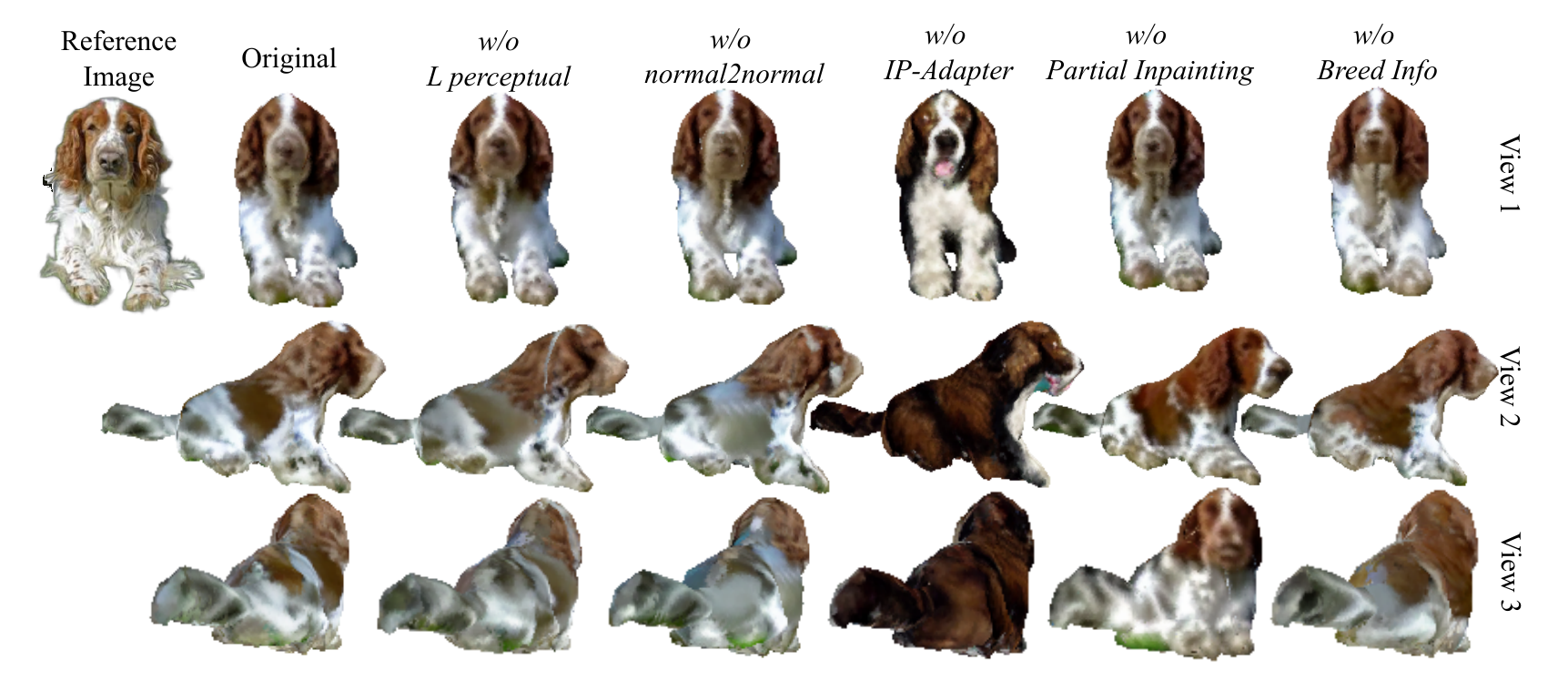}
    \caption{\textbf{Ablation study on key components.} Col.~1 (full pipeline) 
    produces appearance consistent with the reference across all viewpoints. 
    Removing perceptual loss or normal enhancement (Cols.~2--3) introduces 
    subtle geometric and texture degradation. Removing IP-Adapter (Col.~4) 
    causes visible color shift and loss of breed-specific patterns. 
    Removing partial inpainting (Col.~5) produces multi-view inconsistencies. 
    Breed conditioning (Col.~6) sharpens facial feature synthesis.}
    \label{fig:ablation}
\end{figure}

Figure~\ref{fig:ablation} and Table~\ref{tab:results2} evaluate the impact of key components using LPIPS~\cite{zhang2018perceptual} and DreamSim~\cite{fu2023dreamsim}, which measure texture fidelity and perceptual similarity to the input.  

\textbf{Perceptual normal loss.} Removing perceptual losses slightly degrades texture fidelity (LPIPS: 0.2495~$\to$~0.2641) and perceptual similarity (DreamSim: 0.1751~$\to$~0.1821), confirming that perceptually-aligned coarse geometry benefits subsequent SDF refinement and texturing.  

\textbf{Normal enhancement.} Omitting the diffusion-based normal enhancer consistently lowers both metrics (LPIPS: 0.2495~$\to$~0.2661, DreamSim: 0.1751~$\to$~0.1955), showing that high-frequency surface details are essential for realistic appearance across views.  

\textbf{IP-Adapter style control.} Removing IP-Adapter causes the largest drop (LPIPS: 0.2495~$\to$~0.2713, DreamSim: 0.1751~$\to$~0.2546), highlighting its crucial role in maintaining identity-consistent textures, especially in occluded regions.  

\textbf{Breed conditioning.} Excluding breed information moderately reduces perceptual similarity (DreamSim: 0.1751~$\to$~0.1949) with a smaller effect on LPIPS (0.2495~$\to$~0.2570), suggesting breed priors mainly guide high-level identity rather than local texture details.  

Overall, all components contribute meaningfully, with IP-Adapter providing the largest gain, consistent with identity preservation as the primary objective of single-image canine reconstruction.

\subsection{Limitations and Future Work}
\begin{figure}[h!]  
    \centering
    \includegraphics[width=\linewidth]{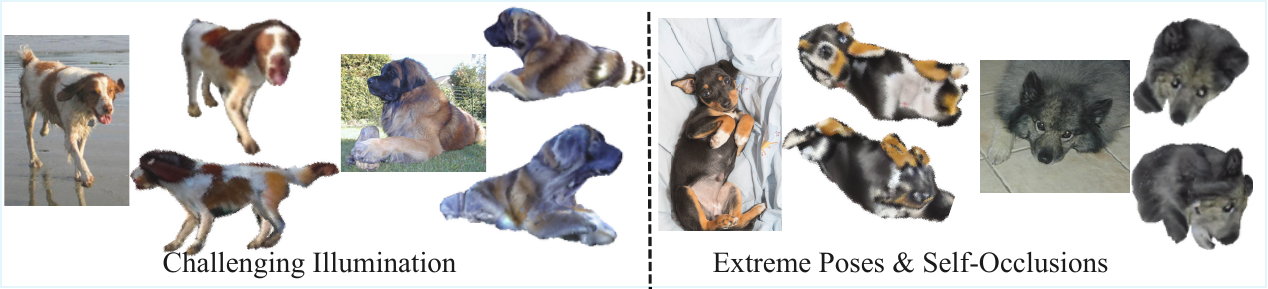}
    \caption{Failure Cases}
    \label{fig:failures}
\end{figure}

\textbf{Inference time.} Our optimization-based pipeline requires ~6 minutes per reconstruction, much slower than feed-forward methods. Future work could explore hybrid approaches combining fast prediction with targeted refinement, or distillation to transfer the multi-stage pipeline into efficient feed-forward networks while retaining style-controlled texture consistency.

\textbf{Domain specificity.} DogWeave relies on BITE for parametric dog initialization. Extending to other quadrupeds (cats, horses, wildlife) would require corresponding parametric priors, e.g., AniMer~\cite{lyu2025animer,lyu2025animerunifiedposeshape}. Nonetheless, our demonstrated benefits of breed conditioning, style control, and monocular perceptual geometry guidance suggest these techniques could generalize effectively under similar dataset conditions.

\textbf{Baked lighting and shadows.} Style control transfers the full appearance of the input image—including lighting and shading—onto the 3D surface (Figure~\ref{fig:failures}, left), producing photorealistic textures but baking illumination into the albedo. While albedo-only supervision could help, IP-Adapter often interprets it as a cartoonish style. Future work may leverage relighting-aware diffusion models to disentangle shading from material appearance without sacrificing realism.

\textbf{Extreme poses and self-occlusion.} Highly occluded poses (e.g., tightly curled bodies) occasionally result in incomplete or inconsistent textures (Figure~\ref{fig:failures}, right). In these cases, geometry-conditioned inpainting struggles because both geometric and appearance cues are ambiguous, limiting the diffusion model's ability to hallucinate plausible textures.

Despite these limitations, DogWeave shows that combining explicit geometry with style-controlled diffusion enables photorealistic reconstruction of shape-ambiguous quadrupeds from single images, capturing diverse appearances and poses, with strong potential to generalize to other animal species given suitable parametric priors and training data.

\section{Conclusion}

We introduced DogWeave, a method for high-fidelity single-image 3D dog reconstruction that achieves photorealistic appearance while maintaining individual identity across viewpoints. Our pipeline integrates template-based initialization, diffusion-enhanced normal refinement, and style-controlled sequential texturing. We demonstrate that enforcing strong texture-geometry correspondence through normal-conditioned diffusion, combined with IP-Adapter style control, significantly improves perceived realism and enables detailed reconstruction of articulated animals with diverse appearances. Extensive evaluation shows DogWeave outperforms existing methods in both identity preservation and photorealism, achieving state-of-the-art results without requiring 3D supervision. Beyond dogs, our framework of combining parametric priors with diffusion-based refinement provides a path toward general single-image reconstruction of deformable, shape-ambiguous objects where appearance and geometry are tightly coupled.

\bibliographystyle{splncs04}
\bibliography{main}
\end{document}